\documentclass[10pt,twocolumn,letterpaper]{article}

\usepackage{cvpr}              
\newcommand{\qisun}[1]{}
\newcommand{\yoda}[1]{}
\newcommand{\warning}[1]{}
\newcommand{\note}[1]{}
\newcommand{\yueyu}[1]{}
\newcommand{\nothing}[1]{}
\newcommand{\modelName}{P2ENet\xspace}
\newcommand{\RR}[1]{\textcolor{black}{#1}}

\usepackage[dvipsnames]{xcolor}

\definecolor{cvprblue}{rgb}{0.21,0.49,0.74}
\usepackage[pagebackref,breaklinks,colorlinks,citecolor=cvprblue]{hyperref}

\usepackage{color}
\usepackage{ifthen}
\usepackage{float}
\usepackage{alltt}
\usepackage{newlfont} 
\usepackage{floatflt}
\usepackage{wrapfig}
\usepackage{fixltx2e}
\usepackage{subcaption}
\usepackage{multirow}
\usepackage{booktabs}

\usepackage{algorithmic}
\usepackage[export]{adjustbox}
\usepackage{mathtools, cuted}
\usepackage{CJKutf8} 

\usepackage[ruled]{algorithm2e} 
\usepackage{enumitem}

\def\paperID{15} 
\def\confName{CVPR}
\def\confYear{2024}

\title{Low Latency Point Cloud Rendering with Learned Splatting}

\author{Yueyu Hu$^{1}$ \quad Ran Gong$^{2}$ \quad Qi Sun$^{1}$ \quad Yao Wang$^{1}$\\
$^{1}$Tandon School of Engineering, New York University \quad $^{2}$Tsinghua University\\
{\tt\small \{yyhu, qisun, yaowang\}@nyu.edu} \quad {\tt\small rangong41@gmail.com} \\ 
}

\begin{document}

\twocolumn[{
    \renewcommand\twocolumn[1][]{#1}
    \maketitle
    \begin{center}
        \centering
        \includegraphics[width=0.9\textwidth]{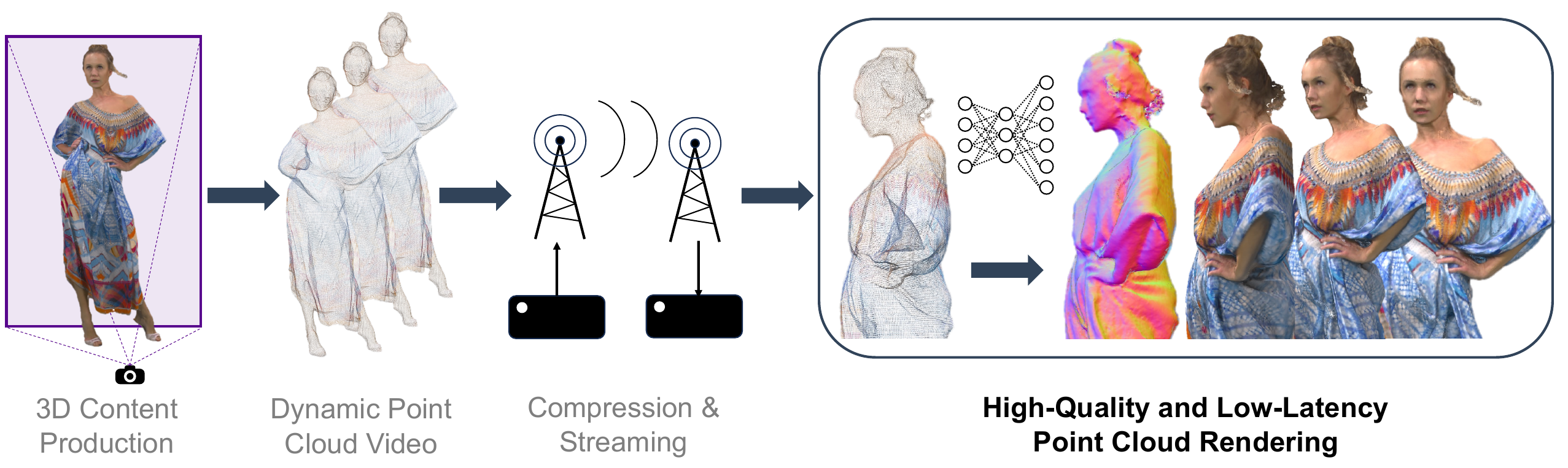}
        \label{fig:teaser}
    \end{center}
}]

\begin{abstract}
    Point cloud is a critical 3D  representation with many emerging applications. Because of the point sparsity and irregularity, high-quality rendering of point clouds is challenging and often requires complex computations to recover the continuous surface representation. On the other hand, to avoid visual discomfort, the motion-to-photon latency has to be very short, under 10 ms. Existing rendering solutions lack in either quality or speed. To tackle these challenges, we present a framework that unlocks interactive, free-viewing and high-fidelity point cloud rendering. We train a generic neural network to estimate 3D elliptical Gaussians from arbitrary point clouds and use differentiable surface splatting to render smooth texture and surface normal for arbitrary views. Our approach does not require per-scene optimization, and enable real-time rendering of dynamic point cloud. Experimental results demonstrate the proposed solution enjoys superior visual quality and speed, as well as generalizability to different scene content and robustness to compression artifacts. The code is available at \url{https://github.com/huzi96/gaussian-pcloud-render}.
\end{abstract}    
\section{Introduction}

Point cloud is a versatile 3D representation that can be directly acquired by various sensors such as LiDAR, multi-view, or RGB-D cameras without heavy processing, thus enabling real-time capturing and streaming.  It is also more flexible than polygonal mesh when representing non-manifold geometry. 
These benefits have led to growing deployment of point cloud for applications such as  culture heritage, autonomous driving, immersive visual communications, and VR/AR. 
Its importance is also evidenced by the MPEG point cloud compression standardization activities since 2014 \cite{chen2023introduction,gpccwhitepaper}. 
However, rendering a point cloud given a user's viewpoint is uniquely challenging: Unlike meshes, the point representation does not provide explicit surface information, making the rendering suffer from point sparsity, geometric irregularity, and sensor noise \cite{eisert2023volumetric}. On the other hand, to avoid visual discomfort, experimental studies have demonstrated that the motion-to-photon (MTP) latency should not exceed 10 ms \cite{grzelka2019impact,mania2004perceptual}. These challenges and demands have become the main roadblocks of the aforementioned point-cloud-enabled applications, especially for scenarios where a dynamic point cloud needs to be captured, streamed, and rendered in real time.

Various solutions have been presented to render point clouds as images from arbitrary viewpoints, including point splatting~\cite{zwicker2001surface,pfister2000surfels,bui2018point} and ray tracing~\cite{chang2023pointersect}.
However, existing methods either require heavy computation (e.g., \cite{chang2023pointersect}), far exceeding the MTP threshold (see \Cref{tab:non-voxelized}), 
or generate blurred images that miss details \cite{kazhdan2006poisson} or have holes \cite{pfister2000surfels} (see \Cref{fig:thuman_256}). 
An alternative solution is cloud-based rendering using a powerful server in the cloud \cite{kaplanyan2019deepfovea,wang2022joint}. This, however, requires the total latency including the round-trip transmission between the user and the server, rendering at the server, and (de)compression of the rendered view, to be completed within the MTP threshold, necessitating extremely low-latency communication links. Even when the server is positioned close to the network edge, the typical delay in today's access networks is still prohibitively large to meet the MTP requirement.



We develop a point cloud rendering framework that enables high visual quality, six degrees of freedom (6-DoF) viewing, and satisfies the MTP requirement with consumer-grade   computers at the user end.
Our solution does not require per-scene neural network training nor heavy surface reconstruction computation.
%
Our method leverages the  3D surface representation using 3D elliptical Gaussians and the corresponding differentiable 3D splatting renderer \cite{kerbl20233d}. We train a light-weight 3D sparse convolutional neural network called Point-to-Ellipsoid (\modelName) to transfer each point in the colored point cloud into an ellipsoid.
The ellipsoids are then splatted to render a frame at the most current viewpoint.  The differentiable renderer enables us to train the \modelName to optimize the rendering quality. The 3D Gaussian representation enables high-quality rendering.
The P2ENet also derives a normal vector with each Gaussian, enables the generation of a normal map beyond a rendered image, and, therefore, unlocks practical applications such as relighting and meshing.

We evaluate our method with both high- and low-density point clouds, as well as a variety of scenes of both dynamic human in actions and outdoor objects.
Experimental results show that our method can render high-quality and hole-less images faster than 100 FPS after an initial delay of less than 30 ms. It is also robust to point cloud capturing and compression noise. Given the affordability of RGB-D capturing devices, we hope the research enables the high quality real-time streaming and rendering of live captured 3D scenes from remote studios, and interactive VR/AR applications among multiple remote participants.   
%
To this aim, we will release our code to the public upon acceptance.  
In summary, we make the following main contributions:
\begin{itemize}[leftmargin=*]
    \item a generalizable neural network \modelName that transforms point clouds to 3D Gaussian representations without per-scene training;
    \item an end-to-end framework for low-latency and high-fidelity point cloud rendering;
    \item the representation enabled high-quality normal maps for practical applications such as meshing and relighting;
\end{itemize}

\section{Related Work}
\label{sec:prior}

\subsection{Point Cloud Rasterization}

Rasterization is a common and efficient method for rendering. However, since points do not naturally have spatial dimensions, they are converted to oriented disk~\cite{pfister2000surfels} or 3D Gaussians~\cite{zwicker2001surface}, which are then  rasterized to pixels. Generally, with screen-space alpha blending by the projected Gaussian kernels, the surface splatting methods~\cite{zhang2007deferred,botsch2004phong,zwicker2004perspective} render smoother surface geometries and visually pleasing texture when the points are dense (i.e. the surface is  adequately sampled). Although the initially proposed 3D Gaussian representation in \cite{zwicker2001surface} is general and can represent an arbitrary tilted ellipsoid through a non-diagonal covariance matrix, earlier methods considered only isotropic Gaussians (i.e. a diagonal covariance matrix with equal diagonal elements).  The variance of the isotropc Gaussian kernel was either set to a global constant and  determined by global point density, or spatially varying depending on the local covariance of point coordinates. To ensure that points in the front surface are correctly occluding points in the back surface, a pixel-level visibility check is usually required~\cite{zhang2007deferred}. The later work in \cite{childs2012auto} proposed an approach to estimate elliptical parameters from a point cloud and further demonstrated that this kind of method can be used for real-time rendering of point clouds. However, due to the diversity in point clouds and existence of capturing and quantization noise, it is hard to determine the optimal variance or elliptical parameters, leading to either holes in the rendered image or blurry texture. This motivates our idea of using a neural network to estimate the optimal elliptical parameters and the displacement for each point by analyzing the local and global structure of the colored point cloud.

\subsection{Learning-based Renderer}


Neural networks are capable of learning complex mapping from inputs to outputs. With a differentiable renderer~\cite{aliev2020neural,yifan2019differentiable,mildenhall2021nerf}, neural networks can be end-to-end trained with supervision on the rendered images. 
Recent works leveraging a differentiable version of the 3D Gaussian splatting renderer have demonstrated capability of inverse rendering, geometry editing~\cite{yifan2019differentiable}, and novel-view synthesis from multi-view images~\cite{ruckert2022adop,kerbl20233d,aliev2020neural}. Pursuing a different direction, \cite{chang2023pointersect} presents a point cloud renderer based on a differentiable ray tracer. By introducing a transformer to estimate the intersection between a camera ray and a local set of points within a cylinder of the ray, the method achieves state-of-the-art rendering quality and enables relighting by simultaneously predicting the surface normal at the intersection point. However, the ray-tracing process, which involves inference of a complex transformer when shading every pixel, is too slow for real-time rendering.

\RR{
Our work is inspired by 3D Gaussian Splatting (3DGS)~\cite{kerbl20233d} which demonstrates that 3D Gaussians can be used to represent smooth surface through per-scene optimization. However, our research is fundamentally orthogonal to the application scenarios of 3DGS and Neural Point-Based Graphics (NBPG)~\cite{aliev2020neural}, which focus on novel view synthesis from multiview images and rely on these images as inputs to generate the point cloud. In contrast, our work is concerned with applications when only pre-captured point cloud data are available and need to be streamed and rendered in real time. Furthermore, 3DGS requires per scene optimization for each individual frame. Therefore, 3DGS cannot support our target application. In comparison, once trained, our model can be used for real-time rendering of various point cloud datasets with not only humans but also natural scenes.
}


\subsection{Alternative Image-Based 3D Representation}
Beyond explicit representation, such as point cloud or mesh, a 3D environment can also be represented in the image space, such as panoramas, light fields, or unstructured multi-view images.
Transforming those data to a given viewpoint requires a view synthesis approach by tackling dis-occlusion and relighting challenges. Example solutions include multi-layer image inpainting \cite{lin2020deep,serrano2019motion}, or neural radiance fields \cite{mildenhall2021nerf,kerbl20233d}.
As an orthogonal scope of work, we focus on rendering the 3D point cloud data with high quality and speed. 
\section{Method}
\label{sec:method}

\begin{figure}
    \centering
    \includegraphics[width=1\linewidth]{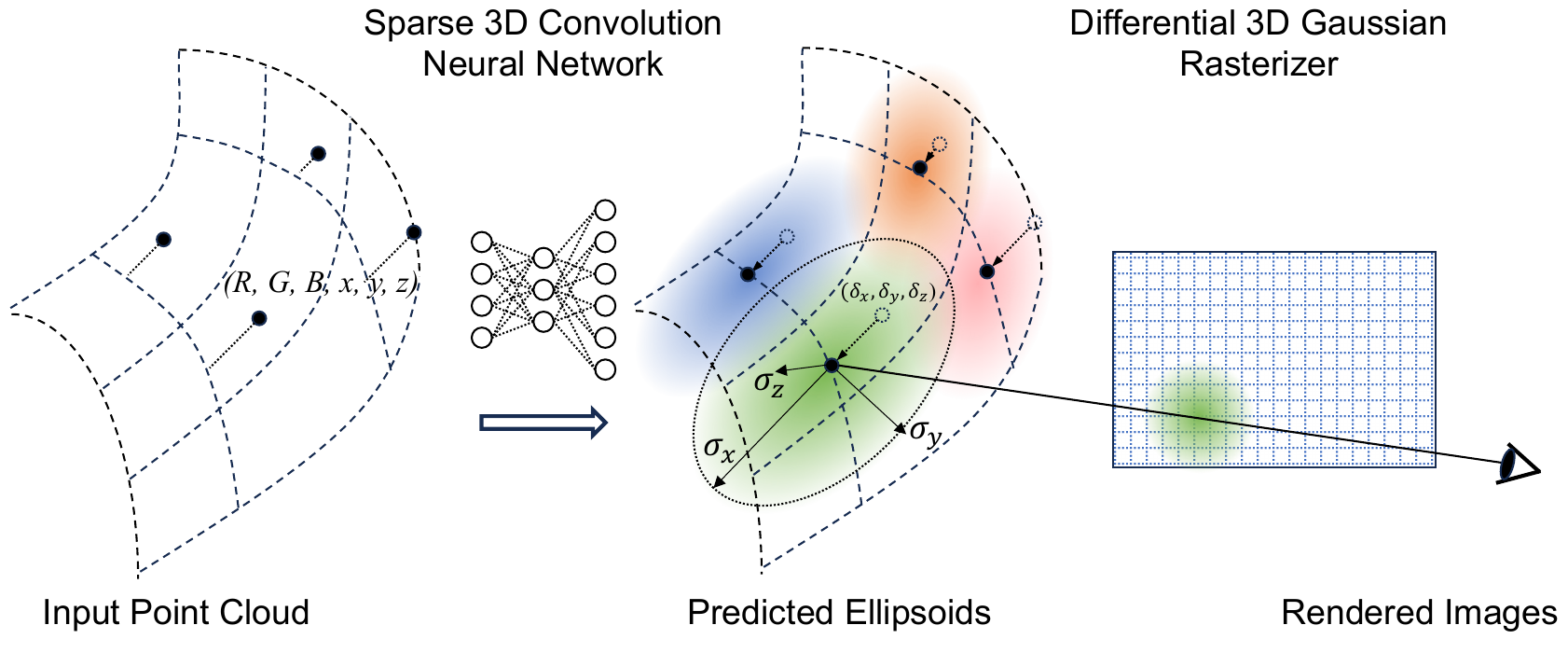}
    \caption{Rendering by estimating elliptical parameters from point cloud using sparse 3D convolutional neural network.}
    \label{fig:main_idea}
    \vspace{-2mm}
\end{figure}

\subsection{Adaptive Surface Splatting Using Elliptical Gaussians}

%
We are given color points sampled from the visible surface in a 3D scene. Each point consists of a 3D coordinate $\mathbf{p} = (x, y, z)$ and an RGB color $\mathbf{c} = (r, g, b)$.
The collection of the points form a frame of 3D point cloud $P =\{(\mathbf{p}_i, \mathbf{c}_i)| i = 1, \cdots, N\}$, and multiple frames form a point cloud video that captures a dynamic scene. These point clouds are usually captured by RGB-D cameras, LiDAR, or reconstructed from multi-view images, and are usually preprocessed to remove noise and outliers.


Given the point cloud, our goal is to render the 3D scene into 2D images from arbitrary views. We mainly address two technical problems. First, points are originally without spatial dimensions. To render smooth textured surface from different views out of points, we need to convert zero-dimensional points into 3D primitives with spatial volumes. Second, since the point clouds may be unevenly spaced and contain quantization noise introduced in compression, we need to adjust the coordinates of the primitives accordingly. Generally, we require the generated primitives to,  1) approximate the surface; 2) avoid visibility leakage; 3) produce smooth texture.

Previous work shows that 3D elliptical Gaussians have the capability to represent a scene and render smooth texture with \RR{hole-free} surface~\cite{kerbl20233d}. Inspired by this, we propose to estimate an ellipsoid (a  3D Gaussian with an arbitrary covariance matrix) from each point, serving as the rendering primitive.
For each input point $((\mathbf{p}, \mathbf{c}))$ in the original point cloud $P$, we estimate a Gaussian center offset~$\delta = (\delta_x, \delta_y, \delta_z)$ and a covariance matrix $\Sigma \in \mathbb{R}^{3\times 3}$, in effect transforming the point into a 3D Gaussian $\mathcal{G}(\mathbf{p} + \delta, \Sigma)$. Specifically, as in \cite{kerbl20233d} the covariance matrix is parameterized as $\Sigma = \mathbf{R}^T \mathbf{S}^T \mathbf{S} \mathbf{R}$, where $\mathbf{R}$ is a rotation matrix and $\mathbf{S} = \text{diag}(\sigma_x, \sigma_y, \sigma_z)$ is a diagonal matrix. This parameterization guarantees $\Sigma$ to be positive semi-definite. The rotation matrix $\mathbf{R}$ is calculated from a quaternion $\mathbf{q} = (q_w, q_x, q_y, q_z)$, which is also estimated by the neural network. The neural network also estimates an opacity value $o$ for each point. Combined with the original color of the point, the final 3D primitive is $<\mathcal{G}(\mathbf{p} + \delta, \Sigma), o, c>$, consisting of 11 parameters: $\delta_x, \delta_y, \delta_z, \sigma_x, \sigma_y, \sigma_z, q_w, q_x, q_y, q_z, o$ to be estimated by the neural network.

We render the 3D Gaussians by splatting them onto the screen space and then rasterize. In practice, we set a threshold of three times the standard deviation of the Gaussian to determine the boundaries of the splats. In the rasterization, one pixel may be covered by multiple splats. Following \cite{kerbl20233d}, We use alpha blending to combine the colors of the splats. Let $x$ be the screen-space coordinate of a pixel and $U = {<\mathcal{G}(\mathbf{p}_k + \delta_k, \Sigma_k), o_k, c_k> | k = 1, \cdots, K}$ be the set of splats that cover the pixel, sorted by the depth (\textit{i.e.} the z coordinate after transforming $\mathbf{p}_k + \delta_k$ into the camera space). The rendered color at $x$ is
\begin{equation}
    \begin{split}
        C &= \sum_{k=1}^K T_k \alpha_k c_k, \text{with }  T_k = \prod_{j=1}^{k-1} (1 - \alpha_j), \\
    \end{split}
    \label{eq:alpha_blending}
\end{equation}
where $c_k$ is the original color of the $k$-th point. We obtain the opacity $\alpha_k$ of the $k$-th splat by projecting the 3D Gaussian onto the screen space and evaluate the Gaussian density at the projected point, weighted by the estimated opacity value $o_k$,
\begin{equation}
    \begin{split}
        \alpha_k &= o_k \cdot e^{-\frac{1}{2} \mathbf{d}_k^T \Sigma_{k(s)}^{-1} \mathbf{d}_k},
    \end{split}
\end{equation}
where $\Sigma_{k(s)}$ denote the screen space covariance matrix and $\mathbf{d}_k$ is the distance between the pixel point and the Gaussian center in the screen space.
It has been shown in \cite{zwicker2001ewa} that, given the viewing transformation $\mathbf{W}$ with camera rotation $\mathbf{T}$ and translation $\mathbf{t}$,
\begin{equation}
    \begin{split}
        \mathbf{W} = \begin{bmatrix}
            \mathbf{T} & \mathbf{t} \\
            \mathbf{0} & 1
        \end{bmatrix},
    \end{split}
\end{equation}
$\Sigma_{k(s)}$ can be approximated as,
\begin{equation}
    \begin{split}
        & \Sigma_{k(s)} = \mathbf{J}\mathbf{T} \Sigma_k \mathbf{T}^T \mathbf{J}^T, \text{with } \mathbf{J} = \begin{bmatrix}
            \frac{f_x}{z}, 0, -\frac{xf_x}{z^2} \\
            0, \frac{f_y}{z}, -\frac{yf_y}{z^2}
        \end{bmatrix},
    \end{split}
\end{equation}
where $f_x$ and $f_y$ are the focal lengths of the camera, and $x, y, z$ are the camera-space coordinate of the projected point.

To enable re-lighting of the rendered images, we also estimate the surface normal of each point $n_k$. The surface normal is estimated by the same neural network as three additional output channels for each point. In the scenario that requires re-lighting, we first render the surface normal to 2D pixels by subsituting $c_k$ in Eq.~(\ref{eq:alpha_blending}) with  $n_k$, and then use the rendered surface normal to calculate the shading. We then conduct late shading with a pixel shader.

\subsection{Neural Elliptical Parameter Estimation}
\label{sec:method:estimator}
\begin{figure*}
    \centering
    \includegraphics[width=\linewidth]{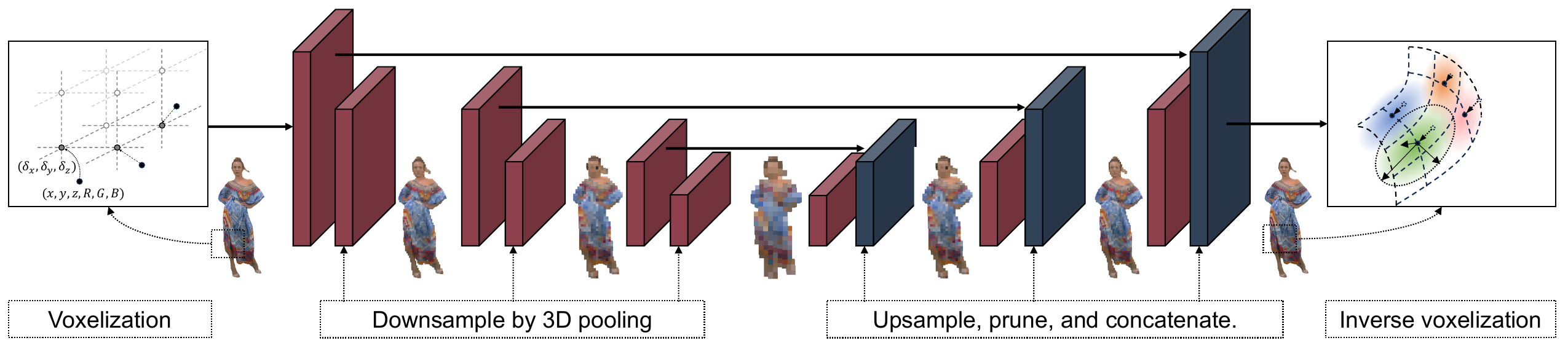}
    \caption{The UNet-like architecture with 3D convolutions. The network takes a point cloud as input and predicts the elliptical parameters and surface normal for each point.}
    \label{fig:net}
    \vspace{-4mm}
\end{figure*}

Unlike the work in \cite{kerbl20233d} which estimates the elliptical parameters from multi-view images using a per-scene optimized neural network, we employ a feed-forward 3D sparse convolutional neural network to estimate the elliptical parameters and surface normal from a given point cloud. The neural network is based on Minkowski Engine~\cite{choy20194d}, which allows efficient 3D voxelized convolutions on sparse data like point clouds. However, the input point clouds may not be originally voxelized. Hence, we need to voxelize the point clouds before feeding them into the neural network. We design a voxelization method that is adaptive to the density of the point cloud and at the same time preserves the information. We first determine a scaling factor such that after scaling, the point cloud has an average density of 1 point per voxel. We then voxelize the scaled point cloud with a voxel size of 1. Since there will be coordinates offset after the voxelization, we also calculate the coordinate residuals, and make it as part of the input feature to the neural network. This ensures that the geometric information is kept. Finally each point we feed into the neural network has the attributes $\{\mathbf{p} = (x, y, z)$, $\mathbf{c} = (r, g, b)$, and $\mathbf{r} = (\delta_x, \delta_y, \delta_z)\}$, namely the absolute coordinates, the color, and the coordinate residuals, respectively.

Approximating the underlying surface represented by the point cloud requires local and global information of the point cloud. To learn the global and local features of the point cloud, we use a UNet-like architecture with 3D convolutions, where the downsampling in the encoder allows the neural network to extract global features from points that are far away from the center point. The network architecture is shown in Fig.~\ref{fig:net}. We use the Minkowski Engine pooling method to downsample the sparse voxel grid. It simply takes every $2 \times 2 \times 2$  voxels and aggregate to one voxel. The new voxel conceptually lies in the center of the original $2 \times 2 \times 2$ voxels  and has the mean feature of those voxels. If some of these voxels are not  occupied, only non-empty voxels will take part in the average calculation. In our architecture, there are 3 downsampling layers in the encoder, effectively enlarging the voxel size by a factor of $8^3$.

The upsampling in the decoder is done by transposed convolution. In an upsampling layer, each non-empty voxel in the input will generate  $2 \times 2 \times 2$  voxels  in the output by transposed 3D convolution. Because the corresponding voxel grid in the encoder may not be fully occupied, we prune the output voxel grid with the ground truth occupancy of the corresponding  voxel grid in the encoder. This guarantees the same geometry of two sparse voxel grid coming from the transposed convolution and the skip connection from the encoder. Symmetrically, the decoder also has 3 layers, with the last layer having the same set of points as the original point cloud. Finally, for each occupied point in the final layer, the network predicts the set of 3D Gaussian parameters $<\mathcal{G}(\mathbf{p} + \delta, \Sigma), o, c, \mathbf{n}>$ from the features of that point in the decoder using a shared MLP.

One of the advantage of using sparse 3D convolution to process voxelized point cloud is that it allows efficient handling of large point clouds with millions of point on GPU. Alternatives like DGCNN~\cite{wang2019dynamic} or PointNet++~\cite{qi2017pointnetplusplus} involve nearest neighbor search, which is memory and time consuming on large scale point clouds. In our experiments, we are able to process high density point clouds  in real-time on a single RTX 4090 GPU.

\subsection{Model Training} \label{sec:training}

We train the neural network using the THuman 2.0 dataset~\cite{yu2021function4d}. This dataset includes meshes of 3D captured human subjects. The human subjects are captured in real-time using a multi-view system with three RGB-D cameras. After that, high-quality texture meshes are reconstructed from the capture and provided as assets.

The training of our system requires input point clouds and the ground truth images from arbitrary viewing directions. We synthesize the training pairs from the meshes. The ground truth rendered RGB images and surface normal maps are directly rasterized from the mesh with random cameras. We obtain the input point clouds for training by mixing both the quantized and non-quantized point clouds sampled from the mesh. We normalize the vertices of the training meshes to be within a bounding box of size $2\times 2 \times 2$ centered at the origin $(0, 0, 0)$. We then use the Poisson Disk~\cite{yuksel2015sample} algorithm to sample 800K points from each mesh. To obtain the quantized point clouds, we apply a scaling factor of 512 to the points within the $2\times 2 \times 2$ bounding box, and round the scaled coordinates into 10 bit integers. The quantized point clouds are scaled back to be aligned with the non-quantized ones, together forming the training inputs.

To simulate real-world scenarios where the point cloud may have sparser areas due to capturing limitation or lossy compression, we randomly downsample the point cloud for data augmentation during training. For quantized point coordinate $x$, we get the downsampled version $\hat{x}$ by $\hat{x} = \lfloor\alpha x\rceil / \alpha, =\alpha \in [0.25, 1]$. For non-quantized point clouds with $N$ points, we randomly choose $\beta N$ points out of the original point set, where $\beta \in [0.125, 1]$.

We randomly place virtual cameras around the scene and calculate the following loss function between the ground truth rasterized RGB images $I$ and normal maps $\mathbf{n}$ generated from the mesh data, with the differential spaltting results $\hat{I}$ and $\hat{\mathbf{n}}$, as,
\nothing{
\begin{equation}
    \begin{split}
        \mathcal{L} = w_1 ||I - \hat{I}||_1 + w_2 \text{LPIPS}(I, \hat{I}) + w_3 ||\mathbf{n} \times \hat{\mathbf{n}}||_2 +  w_4 ||\mathbf{n} - \hat{\mathbf{n}}||_1,
    \end{split}
\end{equation}
}
\begin{equation}
    \begin{split}
        & \mathcal{L}= w_1 ||I - \hat{I}||_1 + w_2 \mathcal{L}_n, \\
        & \mathcal{L}_n = ||\mathbf{n} \times \hat{\mathbf{n}}||_2 +  w_3 \min \left\{||\mathbf{n} - \hat{\mathbf{n}}||_2, ||\mathbf{n} + \hat{\mathbf{n}}||_2 \right\}.
    \end{split}
\end{equation}
In experiments, we use $w_1 = 1, w_2 = 10$, and $w_3 = 0.1$.

\section{Evaluation}
\label{sec:result}

\subsection{Experimental Settings}
\paragraph{Dataset}
To evaluate our methods on various contents and different point cloud qualities, we conduct the experiments with the following datasets:
\begin{itemize}[leftmargin=*]
    \item \textbf{THuman 2.0}~\cite{yu2021function4d}. The THuman 2.0 dataset contains textured meshes of human subjects, captured and reconstructed in real-time by three RGBD camaras. We train on the training split in this dataset with 250 meshes, and evaluate on the testing split with 8 unseen subjects.
    \item \textbf{8iVFB}~\cite{dataset8i}. This is the standard testing dataset for MPEG point cloud compression standardization. It contains high-quality dynamic voxelized point cloud videos of human in action, with each point cloud frame containing 700K to 900K points.
    \item \textbf{BlendedMVS}~\cite{yao2020blendedmvs}. This dataset provides textured meshes of outdoor scenes, captured and reconstructed from a multi-view system. We use this dataset to evaluate the generalizability of our method to various content.
    \item \textbf{CWIPC}~\cite{reimat2021cwipc}. This dataset provides real-time captured raw point cloud from a multi-camera setting, specifically for social XR applications. Each frame contains \~1M points. Since the production of the point cloud does not include surface construction, the point clouds have discontinuous surface and inaccurate point coordinates. We evaluate on this dataset to show the capability of method to be used for live streaming. 
\end{itemize}

\vspace{-3mm}
\paragraph{Baselines for Comparison}
The main benefit of our method is to jointly provide speed and fidelity.
We compare with the following methods in terms of quality and latency. Considering the total latency from the point cloud to final images into account, we categorize these methods into two groups. 
The first group includes \textbf{offline} methods targeting at high fidelity without considering speed.
\begin{itemize}[leftmargin=*]
    \item \textbf{Per-scene Optimized Surface Splatting}~\cite{kerbl20233d}. This method builds a collection of 3D Gaussians from multiview images and then rasterize the Gaussians to render arbitrary views. Note that this method requires per-scene optimization with known multi-view images. We generate these multi-view supervision images by rendering 144 images from the mesh with a virtual camera following a spiral trajectory. Because of the per-scene optimization requirement, it cannot be applied for real-time rendering. We recognize that this method is developed for 3D scene reconstruction from  multi-view images and then novel-view generation, rather then for rendering a point cloud. We use it as a benchmark for the best possible quality we may obtain by building 3D Gaussians from a point cloud. 
    \item \textbf{Pointersect}~\cite{chang2023pointersect}. A state-of-the-art point cloud rendering method utilizing a transformer to calculate ray-point-cloud intersection for ray tracing. Despite the good quality, ray tracing is  slow   and hence   cannot be used for real-time rendering. 
    \item \textbf{Poisson surface reconstruction}~\cite{kazhdan2006poisson}. Rendering a point cloud by first reconstructing a water-tight triangle mesh and then rasterize.
    \end{itemize}
The second group includes {\textbf{real-time}} methods that can render within the MTP constraint:
    \begin{itemize}[leftmargin=*]
    \item \textbf{OpenGL}. Each point is converted to a 1 pixel wide square in the screen space and rasterized to pixels. We use the packaged implementation provided in Open3D~\cite{zhou2018open3d}.
    \item \textbf{Surface splatting using the same isotropic Gaussian representing each point}. We use the same splatting renderer as our proposed method but we use a diagonal covariance matrix with isotropic variance for all three axes, determined globally using average point density.
    
\end{itemize}

For datasets that have mesh representations, we  measure the fidelity of the rendered images from the point cloud compared to the ground truth obtained from the original scene mesh, using quality metrics PSNR and MS-SSIM~\cite{wang2003multiscale}. In addition to the original point cloud, we also evaluate these methods in the scenario where the point clouds are lossily compressed for streaming. We use the standard G-PCC~\cite{graziosi2020overview} to compress the point clouds at different bit-rate.
For other datasets without meshes, we present visual comparisons for selected point clouds.

\subsection{Fidelity, Latency, and Robustness to Compression Artifacts}

\begin{table*}[htbp]
    \centering
    \caption{Average PSNR (dB) and MS-SSIM scores of rendered views from original point clouds in the THuman 2.0 testing set. We also report preprocessing (P) and rendering (rasterization or ray-tracing) (R) latencies.}
    \footnotesize
    \begin{tabular}{c|cccc|cccc}
        \toprule
        \multirow{2}{*}{Method} & \multicolumn{4}{c|}{Compact (280K, quantized)} & \multicolumn{4}{c}{High-quality (800K, non-quantized)} \\
        & PSNR $\uparrow$ & MS-SSIM $\uparrow$  & Latency (P) $\downarrow$  & Latency (R) $\downarrow$  & PSNR & MS-SSIM  & Latency (P) $\downarrow$  & Latency (R) $\downarrow$ \\
        \midrule
        Pointersect& 30.8 & 0.9926 & $<$1 ms  & 1 s & 33.7 & \textbf{0.9954} & $<$ 1ms & 1 s \\
        Poisson    & 28.5 & 0.9739 & 19 s &  2 ms & 28.7     & 0.9748 & 40s & 2 ms\\

        Global 
        Parameter  & 28.6 & 0.9863 & $<$1 ms (190 ms)\footnotemark[1] & $<$ 1 ms & 30.3 & 0.9924 & $<$1 ms (570 ms)\footnotemark[1] & $<$ 1ms \\
        OpenGL     & 29.2 & 0.9903 & 2 ms (190 ms)\footnotemark[1] & 2 ms\footnotemark[2] & 29.2 & 0.9903 & 3 ms (570 ms)\footnotemark[1] & 3 ms\footnotemark[2]  \\
        \midrule
        Ours & \textbf{33.8} & \textbf{0.9952}  & 27 ms& $<$1 ms & \textbf{34.1} & \textbf{0.9954} & 70 ms & 1 ms \\
        \midrule
        Per-Scene
         3D GS      & \underline{34.5} & 0.9946 & $>$ 5 min & $<$ 1 ms & \underline{34.2} & 0.9942 & $>$ 7min & $<$ 1 ms \\
        \bottomrule
    \end{tabular}
    \label{tab:non-voxelized}
    \vspace{-2mm}
\end{table*}


We first evaluate the method's capability of rendering high-quality point clouds sampled from a smooth surface. From  the testing split of the THuman 2.0 dataset, we create  two categories of point cloud for  each mesh: 1) \textbf{High-quality} non-quantized point cloud sampled from the mesh surface using  Poisson Disk ~\cite{yuksel2015sample}. We sample 800K points from each individual mesh asset; 2) \textbf{Compact} point cloud uniformly sampled from the mesh, with 280K points on average, with coordinates quantized to a 10-bit depth. (see more details in Sec.~\ref{sec:training}). Quantization and down-sampling of points are common tools used for compression of point cloud data codec~\cite{graziosi2020overview}, necessary for efficient point cloud data storage and delivery.

We render point clouds from 12 different viewing angles, forming a circle trajectory surrounding the subject. We report the PSNR and MS-SSIM between the rendered views and the ground truth views rasterized from the meshes. For this evaluation, we render images with a $512 \times 512$ resolution.

Usually a point cloud rendering method consists of a two-stage procedure: 1) constructing primitives from the point cloud (\textit{preprocess}), and 2) render from a camera view by rasterization. 
In  real-time video streaming applications, it is important for the method to finish preprocessing faster than the content frame rate, in addition to have a rendering time that is within the MTP threshold. 
Hence we also compare these two latencies among different methods. The latencies are measured on a computer with an Intel i7-9700K CPU and an NVIDIA RTX 4090 GPU.

\nothing{As shown in \Cref{tab:non-voxelized}, our method demonstrates competitive \qisun{Commonly, to make any statement, we want to provide numbers (say the mean value is higher/lower), if the direct mean/std comparison doesn't work out, we can do statistical analysis such as ANOVA showing at least another condition is not significantly better than our approach.}
rendering quality to offline methods (Pointersect~\cite{chang2023pointersect}, Poisson Surface Reconstruction (Poisson)~\cite{kazhdan2006poisson}, and 3D Gaussian Splatting (3D GS)~\cite{kerbl20233d}) and shows superior quality \qisun{summarizing with stat numbers} than other real-time methods.}

As shown in \Cref{tab:non-voxelized}, our method demonstrates higher rendering quality than all compared methods, with more than 4dB improvement in PSNR than real-time baselines (``Global parameters'' and ``OpenGL''), for both ``high-quality'' and ``compact" data. Our method is also substantially more robust to quantization noise and reduced point density, compared to all other methods. Whereas the PSNR only reduced from 34.1 dB for the unquantized 800K data to 33.8 dB for the quantized 280K data, some of the baseline methods suffer a reduction of 2 to 3 dB.  
We attribute the robustness of our method to training of the neural network using both unquantized and quantized data at randomized point densities. The model can generate accurate Gaussian parameters even when the point cloud is compressed.

\begin{figure*}[t]
    \centering
    
    \begin{subfigure}{0.15\linewidth}
        \centering
        \includegraphics[width=\linewidth]{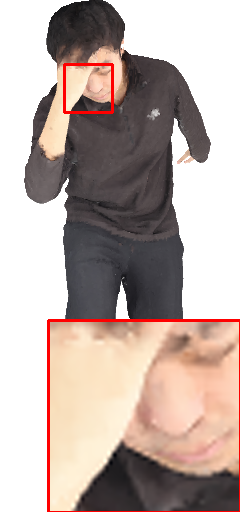}
    \end{subfigure}
    \begin{subfigure}{0.15\linewidth}
        \centering
        \includegraphics[width=\linewidth]{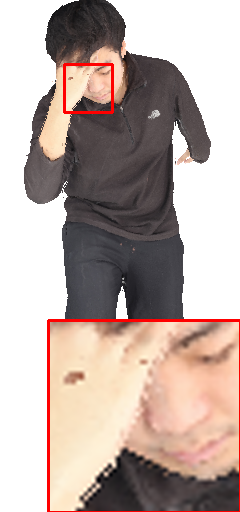}
    \end{subfigure}
    \begin{subfigure}{0.15\linewidth}
        \centering
        \includegraphics[width=\linewidth]{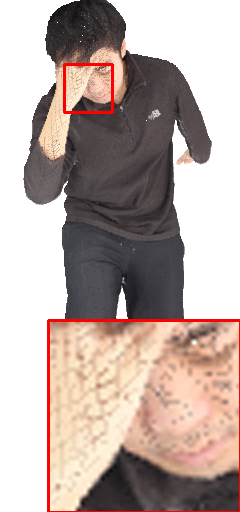}
    \end{subfigure}
    \begin{subfigure}{0.15\linewidth}
        \centering
        \includegraphics[width=\linewidth]{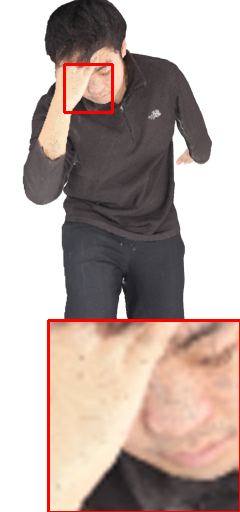}
    \end{subfigure}
    \begin{subfigure}{0.15\linewidth}
        \centering
        \includegraphics[width=\linewidth]{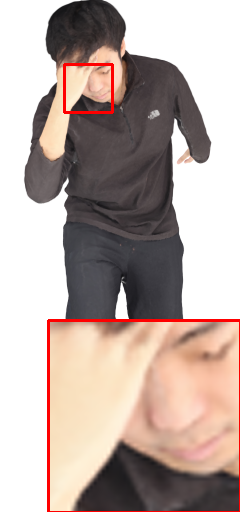}
    \end{subfigure}
    \begin{subfigure}{0.15\linewidth}
        \centering
        \includegraphics[width=\linewidth]{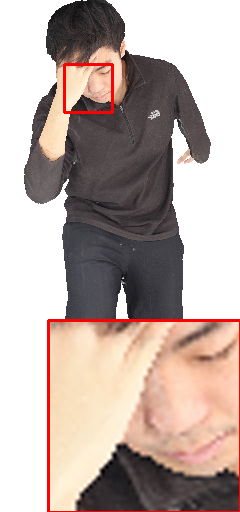}
    \end{subfigure}

    \begin{subfigure}{0.15\linewidth}
        \centering
        \includegraphics[width=\linewidth]{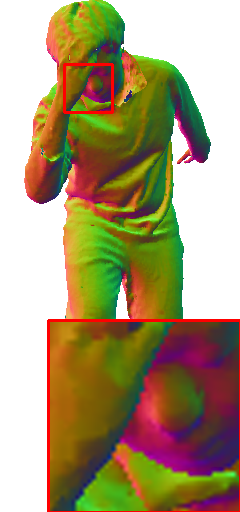}
    \end{subfigure}
    \begin{subfigure}{0.15\linewidth}
        \centering
        \includegraphics[width=\linewidth]{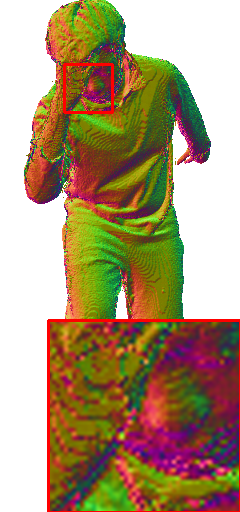}
    \end{subfigure}
    \begin{subfigure}{0.15\linewidth}
        \centering
        \includegraphics[width=\linewidth]{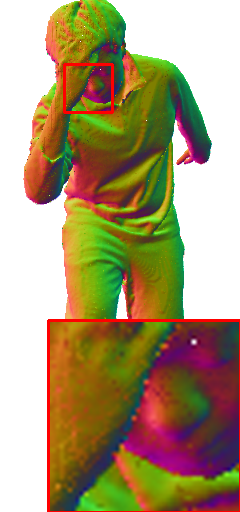}
    \end{subfigure}
    \begin{subfigure}{0.15\linewidth}
        \centering
        \includegraphics[width=\linewidth]{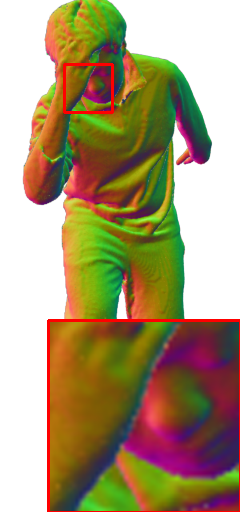}
    \end{subfigure}
    \begin{subfigure}{0.15\linewidth}
        \centering
        \includegraphics[width=\linewidth]{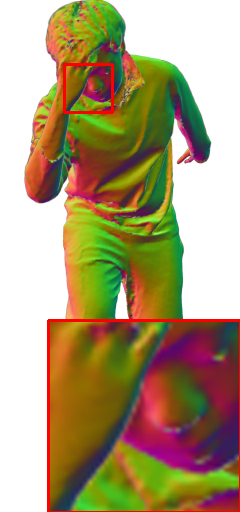}
    \end{subfigure}
    \begin{subfigure}{0.15\linewidth}
        \centering
        \includegraphics[width=\linewidth]{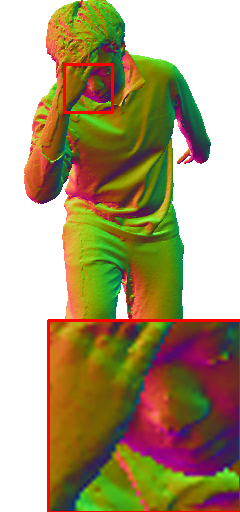}
    \end{subfigure}

    \begin{subfigure}{0.15\linewidth}
        \centering
        Poisson
    \end{subfigure}
    \begin{subfigure}{0.15\linewidth}
        \centering
        Pointersect
    \end{subfigure}
    \begin{subfigure}{0.15\linewidth}
        \centering
        OpenGL 
    \end{subfigure}
    \begin{subfigure}{0.15\linewidth}
        \centering
        Global Parameter
    \end{subfigure}
    \begin{subfigure}{0.15\linewidth}
        \centering
        Ours
    \end{subfigure}
    \begin{subfigure}{0.15\linewidth}
        \centering
        Mesh
    \end{subfigure}

    \caption{Rendering results in the \textit{compact} setting of a point cloud in the Thuman 2.0 Dataset. The first row shows the rendered RGB views. The second row shows the surface normal. The insets visualize local details with $4\times$ zooming.\nothing{\qisun{Do the zoom-in in a more aggressive way to show those micro detail differnces.}}}
    \label{fig:thuman_256}
\end{figure*}

We visualize the rendering results for a sample  compact point cloud in \Cref{fig:thuman_256}. As shown, despite the high rendering latency, Poisson mesh reconstruction still has noticeable blurring artifacts, and Pointersect produces noisy edges due to inaccurate intersection calculation. Among the real-time methods, OpenGL fails to reproduce hole-less surface and leads to visible holes and gaps. 3D Gaussians with global parameters produce inaccurate geometry and noisy edges. Our method generally produces better visual quality than both offline and real-time methods. Please refer to the supplementary material for results with high-quality non-quantized point clouds.

\begin{figure}
    \centering
    \includegraphics[width=0.7\linewidth]{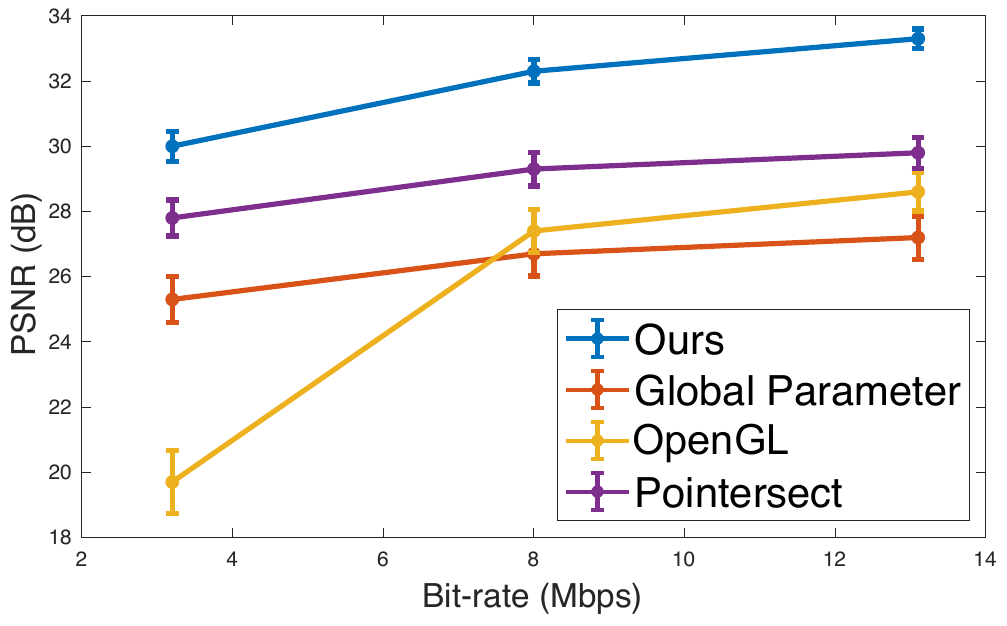}
    \caption{Average PSNR of rendered views from THuman 2.0 \textit{compact} point clouds compressed to different bit-rates by G-PCC.}
    \label{fig:gpcc_psnr}
    \vspace{-6mm}
\end{figure}

\begin{figure*}[t]
    \centering
    \begin{subfigure}{0.19\linewidth}
        \includegraphics[width=\linewidth]{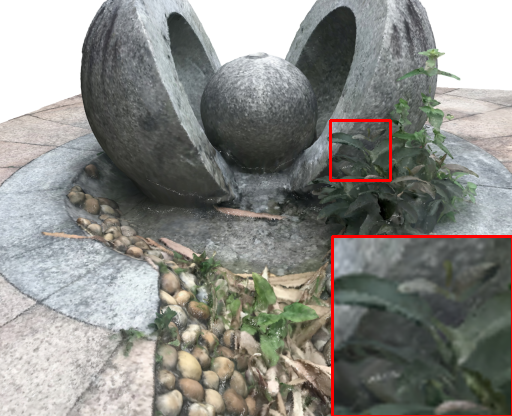}
        \centering
    \end{subfigure}
    \begin{subfigure}{0.19\linewidth}
        \includegraphics[width=\linewidth]{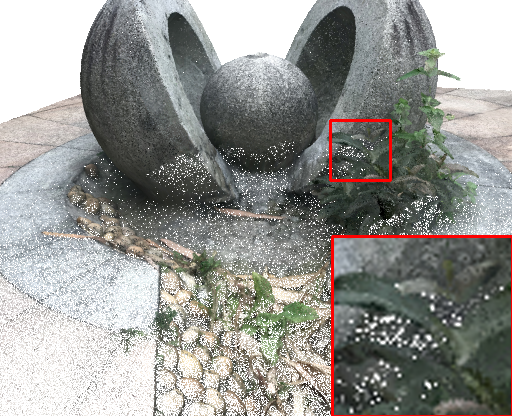}
    \centering
    \end{subfigure}
    \begin{subfigure}{0.19\linewidth}
        \includegraphics[width=\linewidth]{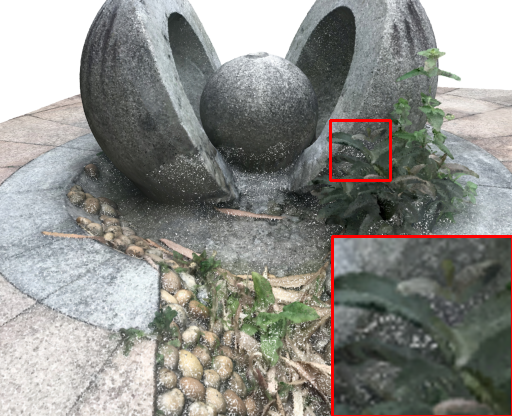}
        \centering
    \end{subfigure}
    \begin{subfigure}{0.19\linewidth}
        \includegraphics[width=\linewidth]{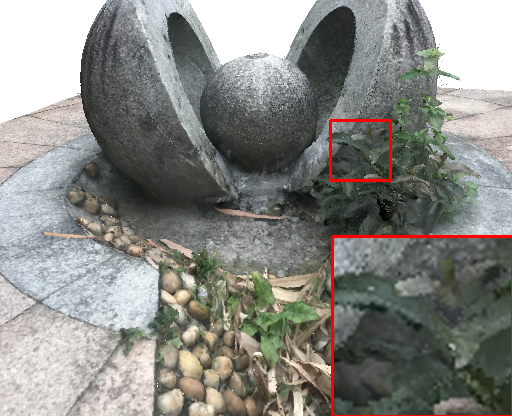}
        \centering
    \end{subfigure}
    \begin{subfigure}{0.19\linewidth}
        \includegraphics[width=\linewidth]{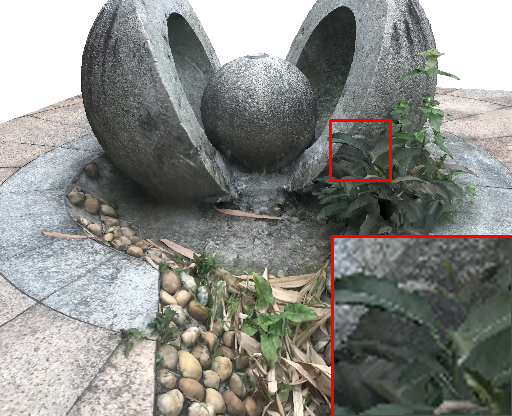}
        \centering
    \end{subfigure}

    \centering
    \begin{subfigure}{0.19\linewidth}
        \includegraphics[width=\linewidth]{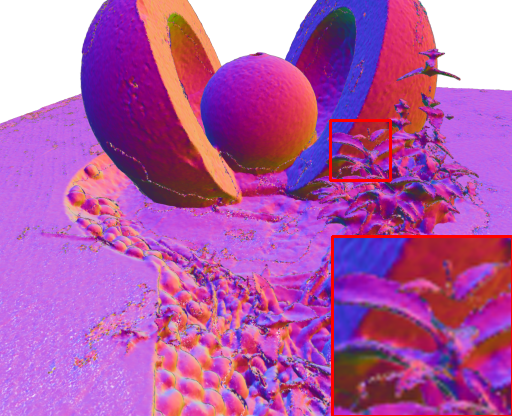}
        \centering
        \scriptsize Ours (23.8 dB, 0.9153)
    \end{subfigure}
    \begin{subfigure}{0.19\linewidth}
        \includegraphics[width=\linewidth]{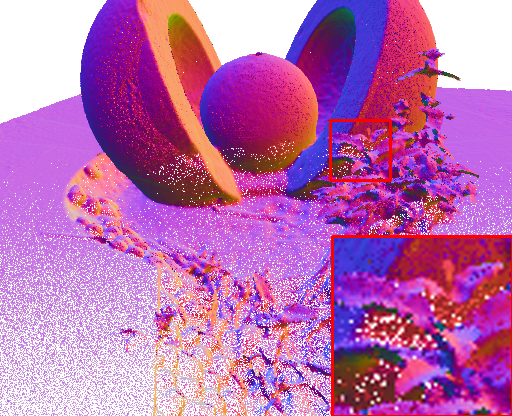}
    \centering
        \scriptsize OpenGL (15.0 dB, 0.7083)
    \end{subfigure}
    \begin{subfigure}{0.19\linewidth}
        \includegraphics[width=\linewidth]{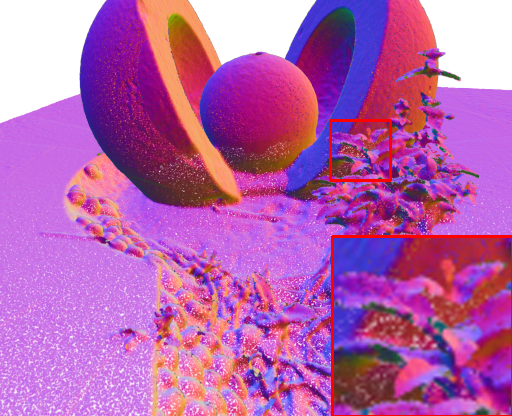}
        \centering
        \scriptsize Global Param. (21.4 dB, 0.8622)
    \end{subfigure}
    \begin{subfigure}{0.19\linewidth}
        \includegraphics[width=\linewidth]{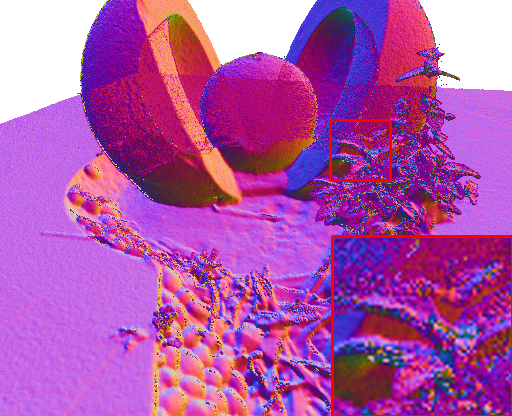}
        \centering
        \scriptsize Pointersect (24.0 dB, 0.9293)
    \end{subfigure}
    \begin{subfigure}{0.19\linewidth}
        \includegraphics[width=\linewidth]{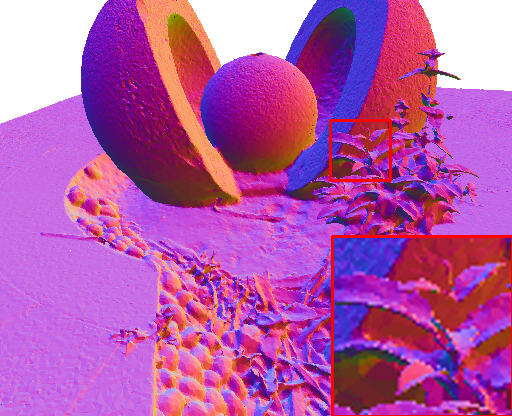}
        \centering
        \scriptsize Ground Truth Mesh
    \end{subfigure}
    
    \caption{Comparison of rendering results of a point cloud (1.7M points) in the BlendedMVS dataset. Quality metrics are shown in format (PSNR, MS-SSIM) and calculated from the rasterization results of the ground truth mesh.  The insets visualize local details with $3\times$ zooming.}
    \label{fig:blended_1003}
    \vspace{-3mm}
\end{figure*}

Our method also enjoys very fast rendering speed once the Gaussian parameters are inferenced using the \modelName, comparable to the real-time baselines, shorter than the MTP threshold. The preprocessing time (for inferencing the Gaussian parameters) is under 30 ms, which should be acceptable for most applications.
Therefore, our rendering method enables the streaming and free viewing of a 30 fps point cloud video after an initial delay of 30 ms (considering only the preprocessing delay). 
Given the rendering time of under 1 ms, the proposed method enables a display frame rate of more than 100 fps.
Since the rendering resolution is irrelevant to the preprocessing, our method is capable of rendering at even higher resolution at high frame-rate, \textit{e.g.} 1K$\times$1K at 1.1 ms and 2K$\times$2K at 2.0 ms, per frame.
Note that such parallel inference-rendering framework is compatible with most consumer grade computers with an integrated GPU on the CPU chip and a discrete GPU in the graphics card. The discrete GPU can be used for inference, while the integrated  GPU can be used for rendering.

\begin{figure*}[t]
    \centering

    \begin{subfigure}{0.18\linewidth}
    \centering
        \includegraphics[width=\linewidth]{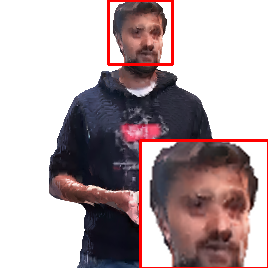}
        Poisson
    \end{subfigure}
    \begin{subfigure}{0.18\linewidth}
    \centering
        \includegraphics[width=\linewidth]{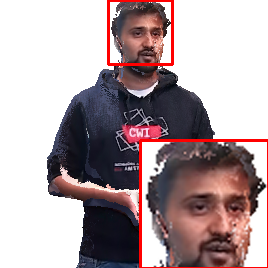}
        Pointersect
        \end{subfigure}
    \begin{subfigure}{0.18\linewidth}
    \centering
        \includegraphics[width=\linewidth]{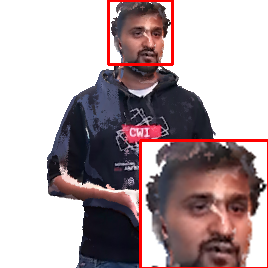}
        OpenGL
    \end{subfigure}
    \begin{subfigure}{0.18\linewidth}
    \centering
        \includegraphics[width=\linewidth]{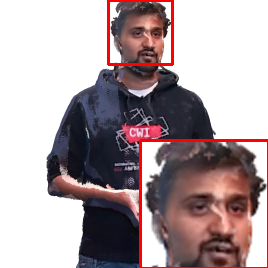}
        Global Parameters
    \end{subfigure}
    \begin{subfigure}{0.18\linewidth}
    \centering
        \includegraphics[width=\linewidth]{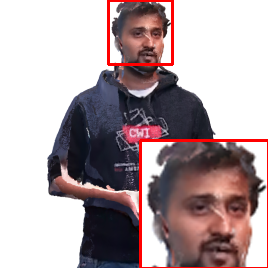}
        Ours
    \end{subfigure}
    
    \caption{Rendering results from noisy raw point clouds in the CWIPC dataset. The insets are with $2\times$ and $3\times$ zooming, respectively.}
    \label{fig:cwi}
    \vspace{-2mm}
\end{figure*}

To more thoroughly evaluate the robustness of the rendering methods to compression,  we use the standard point cloud codec G-PCC \cite{tmc13} to further compress the ``Compact'' versions of the point clouds at different bit-rates, and use different renderers to render the decoded point clouds. Note that G-PCC compresses each frame independently and achieves different rates by rescaling and quantizing the point coordinates (which has the effect of reducing the point density) plus additional color quantization.  As shown in \Cref{fig:gpcc_psnr}, our method consistently achieves higher rendering quality at different bit-rate levels.
%
\nothing{Due to the large data volume of raw point clouds, to efficiently stream volumetric contents over the network, compression is  an essential component in a practical streaming system~\cite{viola2023volumetric}. Therefore, the point cloud renderer in the system must be robust to degradation caused by lossy compression.}
The robustness of our rendering method to  compression, in addition to its fast speed,  makes it more suitable for streaming applications.



\footnotetext[1]{The method does not directly provide surface normal. In the bracket is the time for point cloud normal estimation by Open3D on CPU.}
\footnotetext[2]{Implemented on CPU by Open3D.}

\subsection{Generalizability}


Although our model is trained on the THuman dataset with only high-quality 3D human scans, it generalize to different types of point clouds including outdoor scenes in the BlendedMVS dataset and noisy point clouds in the CWIPC dataset. We show rendering results with the BlendedMVS dataset in \Cref{fig:blended_1003}. Please refer to the supplementary material for results with the CWIPC dataset. Despite the discrepancy in contents and capturing quality between the tested point clouds and training samples, our model still produces comparable quality among the best rendering methods. However, since our training data do not simulate misalignment capturing artifacts and severe noise, it has limitations in handling these quality degradations. We therefore leave it for an aspect of future work.

\section{Limitations and Future Work}
\label{sec:discussion}
\RR{
Our pre-trained 3D sparse CNN (\Cref{sec:method:estimator}) allows us to estimate the elliptical Gaussian attributes from point clouds, without the per-scene training in current view synthesis methods \cite{kerbl20233d,mildenhall2021nerf}. 
It also ensures consistent real-time speed.
However, different scene content types also exhibit varied image quality metrics. For example, the human-body-based point clouds show elevated quality than natural scenes and we observe that our method tends to produce over smoothed images when the point clouds are sparse. Our model also needs to be further finetuned for scenes with a particular irregularity in point density  caused by specific capturing setup for better robustness. 
On the other hand, since our method is optimized to produce high quality renders from each frame of the point cloud video, and does not specifically optimize for temporal consistency, the temporal jittering caused by point cloud capturing can be still observed in the rendered video.}

\RR{In the future, we will invest data augmentation approaches that balances various scene types and noise levels. We also plan to include temporal coherence constraints in model training, and further make the neural network generate denser 3D Gaussians for areas of complex texture, to improve render quality in both spatial and temporal domains.}

\section{Conclusion}
\label{sec:conclusion}
In this paper, we present an end-to-end and learning-based framework that addresses the challenging dilemma between speed and quality in point cloud rendering.
To this end, we leveraged a differentiable, splatting-in-the-loop approach that can generate  fine-grained geometric and textural details through a learnt 3D sparse neural network.
Extensive comparisons with a broad spectrum of datasets and alternative solutions demonstrated the effectiveness of the method.
We hope the research to contribute a new building block for enabling the flexible point cloud as a promising medium toward high-fidelity interactive computer graphics, VR/AR, and immersive visual communications,  while passing the required MTP for user-centric applications.

\section*{Acknowledgement}
This material is based upon work supported by the National Science Foundation under Grant No. 2312839.
{
    \small
    \bibliographystyle{./ieeenat_fullname}
    \bibliography{main}
}

\clearpage
\setcounter{page}{1}
\maketitlesupplementary

\section*{Experiments on Texture Adaptability}

\begin{figure*}
    \centering
    \begin{subfigure}{0.48\linewidth}
        \includegraphics[width=0.8\linewidth]{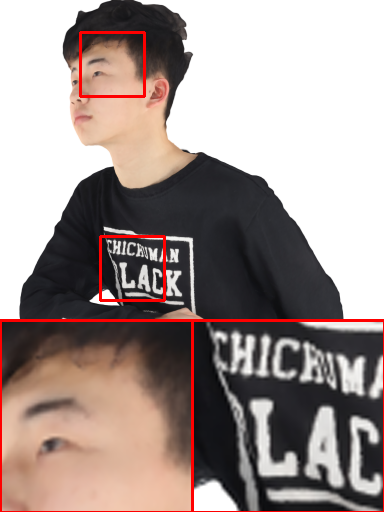}\\
        \centering
        Learned Elliptical Parameters
    \end{subfigure}
    \begin{subfigure}{0.48\linewidth}
        \includegraphics[width=0.8\linewidth]{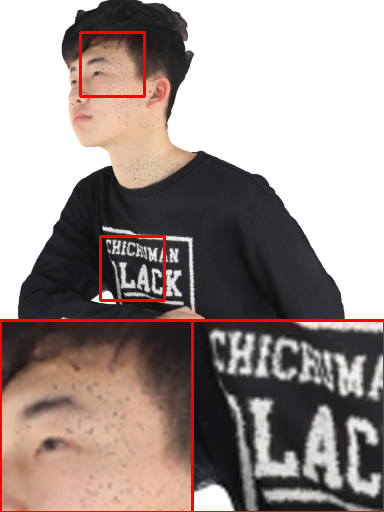}\\
        \centering
        Global $\sigma=1.2 \bar{d}$
    \end{subfigure}

    \begin{subfigure}{0.48\linewidth}
        \includegraphics[width=0.8\linewidth]{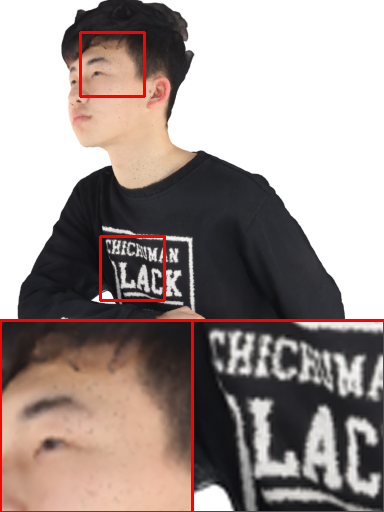}\\
        \centering
        Global $\sigma=1.5 \bar{d}$
    \end{subfigure}
    \begin{subfigure}{0.48\linewidth}
        \includegraphics[width=0.8\linewidth]{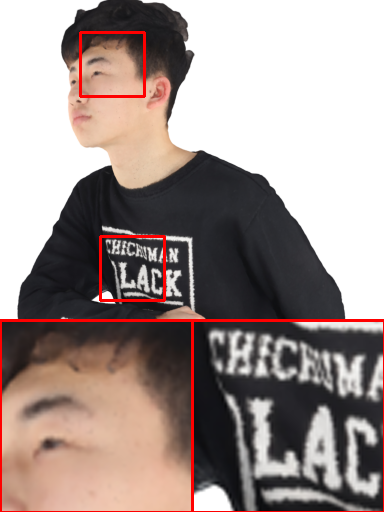}\\
        \centering
        Global $\sigma=1.7 \bar{d}$
    \end{subfigure}
    \caption{Comparison of texture rendering results with our model predicted vs. globally set Gaussian parameters on a quantized point cloud. The insets visualize local details with $3\times$ zooming.}
    \label{fig:texture}
    \vspace{-4mm}
\end{figure*}

In addition to the benefit of rendering more accurate surface contour by allowing 3D Gaussians to have elliptical shapes, since the model also takes point color as input, it has the ability to adapt to local texture edges. We compare the rendering results from learned and globally set Gaussian parameters to elucidate this advantage. For the globally set parameters, we choose three different values of Gaussian standard deviation $\sigma$ relative to $\bar{d}$, the average distance between points in the point cloud. As shown in \Cref{fig:texture}, using a lower global $\sigma$  can lead to visible holes, while setting a larger $\sigma$ result in inaccurate edges. The learned model can instead produce spatially varying 3D Gaussian parameters including center translations to provide clearer edges and smooth surface at the same time.

\section*{Additional Rendering Results}
\label{sec:visual_results}

\begin{figure*}
    \centering

    \begin{subfigure}{0.19\linewidth}
    \centering
        \includegraphics[width=\linewidth]{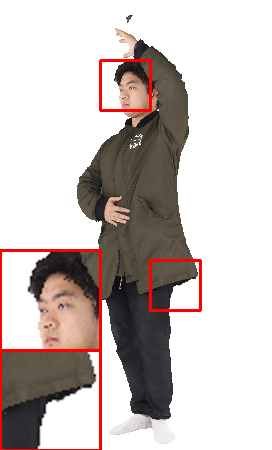}
    \end{subfigure}
    \begin{subfigure}{0.19\linewidth}
      \centering
          \includegraphics[width=\linewidth]{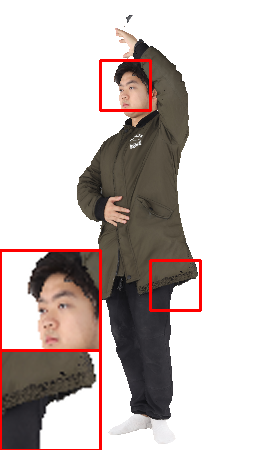}
      \end{subfigure}
      \begin{subfigure}{0.19\linewidth}
      \centering
          \includegraphics[width=\linewidth]{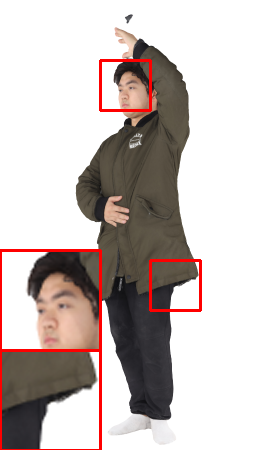}
      \end{subfigure}
      \begin{subfigure}{0.19\linewidth}
      \centering
          \includegraphics[width=\linewidth]{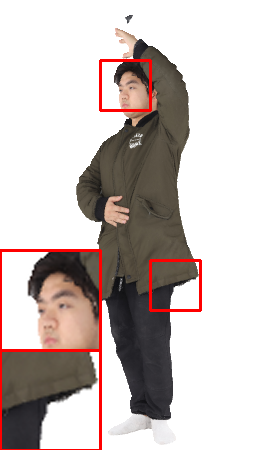}
      \end{subfigure}
      \begin{subfigure}{0.19\linewidth}
      \centering
          \includegraphics[width=\linewidth]{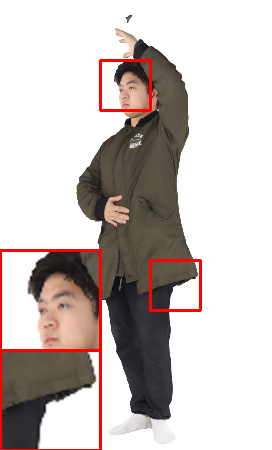}
      \end{subfigure}

    \begin{subfigure}{0.19\linewidth}
    \centering
        \includegraphics[width=\linewidth]{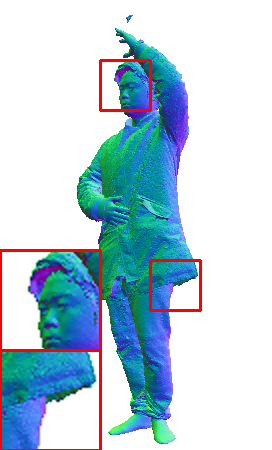}
        Mesh Ground Truth
    \end{subfigure}
    \begin{subfigure}{0.19\linewidth}
    \centering
        \includegraphics[width=\linewidth]{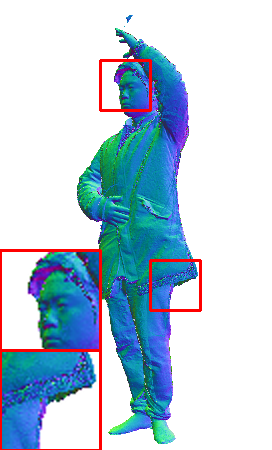}
        Pointersect~\cite{chang2023pointersect}
    \end{subfigure}
    \begin{subfigure}{0.19\linewidth}
    \centering
        \includegraphics[width=\linewidth]{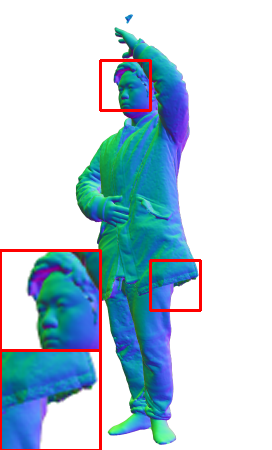}
        Global Parameter
    \end{subfigure}
    \begin{subfigure}{0.19\linewidth}
    \centering
        \includegraphics[width=\linewidth]{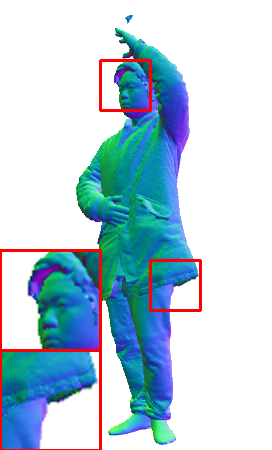}
        OpenGL
    \end{subfigure}
    \begin{subfigure}{0.19\linewidth}
    \centering
        \includegraphics[width=\linewidth]{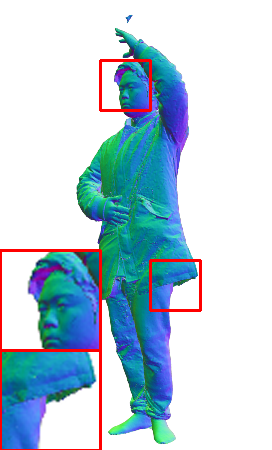}
        Ours
    \end{subfigure}
    
    \caption{Rendering results of high quality point clouds in THuman 2.0 dataset. The insets visualize local details with $2\times$ zooming.}
    \label{fig:0513_800k}
    \vspace{-4mm}
\end{figure*}

\begin{figure*}
  \centering

  \begin{subfigure}{0.19\linewidth}
  \centering
      \includegraphics[width=\linewidth]{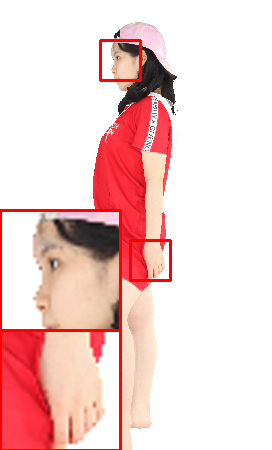}
  \end{subfigure}
  \begin{subfigure}{0.19\linewidth}
    \centering
        \includegraphics[width=\linewidth]{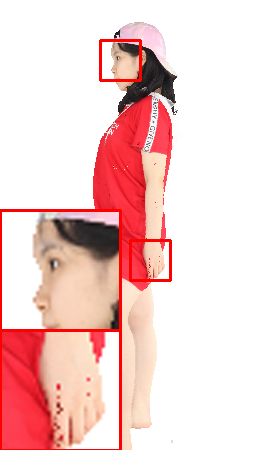}
    \end{subfigure}
    \begin{subfigure}{0.19\linewidth}
    \centering
        \includegraphics[width=\linewidth]{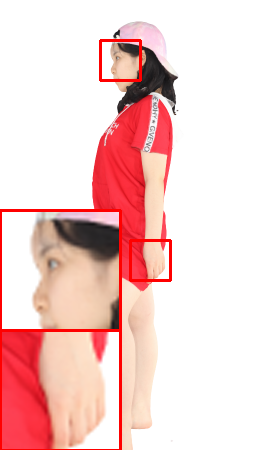}
    \end{subfigure}
    \begin{subfigure}{0.19\linewidth}
    \centering
        \includegraphics[width=\linewidth]{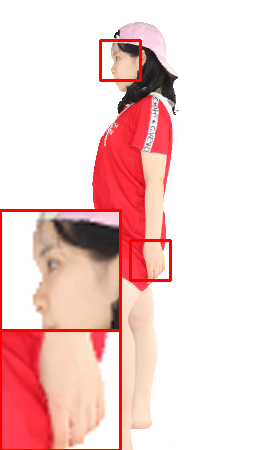}
    \end{subfigure}
    \begin{subfigure}{0.19\linewidth}
    \centering
        \includegraphics[width=\linewidth]{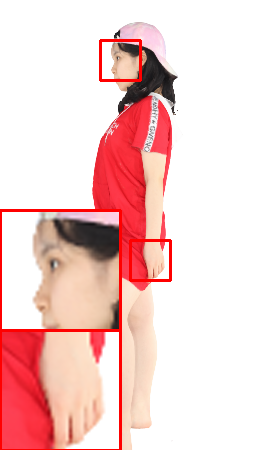}
    \end{subfigure}

  \begin{subfigure}{0.19\linewidth}
  \centering
      \includegraphics[width=\linewidth]{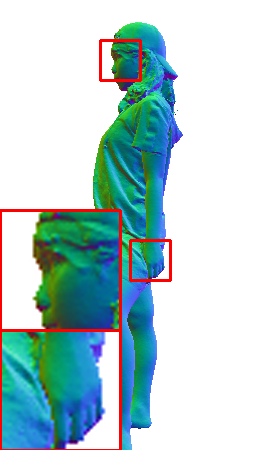}
      Mesh Ground Truth
  \end{subfigure}
  \begin{subfigure}{0.19\linewidth}
  \centering
      \includegraphics[width=\linewidth]{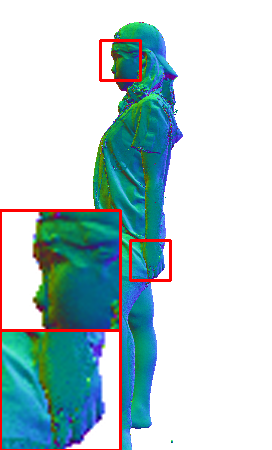}
      Pointersect~\cite{chang2023pointersect}
  \end{subfigure}
  \begin{subfigure}{0.19\linewidth}
  \centering
      \includegraphics[width=\linewidth]{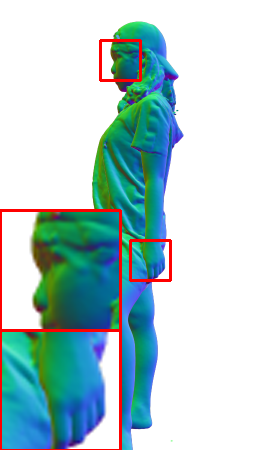}
      Global Parameter
  \end{subfigure}
  \begin{subfigure}{0.19\linewidth}
  \centering
      \includegraphics[width=\linewidth]{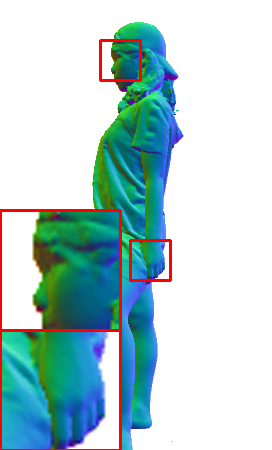}
      OpenGL
  \end{subfigure}
  \begin{subfigure}{0.19\linewidth}
  \centering
      \includegraphics[width=\linewidth]{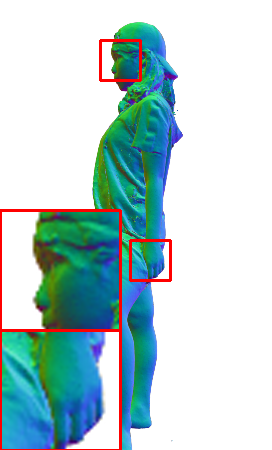}
      Ours
  \end{subfigure}
  
  \caption{Rendering results of high quality point clouds in THuman 2.0 dataset. The insets visualize local details with $3\times$ zooming.}
  \label{fig:0519_800k}
\end{figure*}

\begin{figure*}[t]
  \centering

  \begin{subfigure}{0.3\linewidth}
  \centering
      \includegraphics[width=0.9\linewidth]{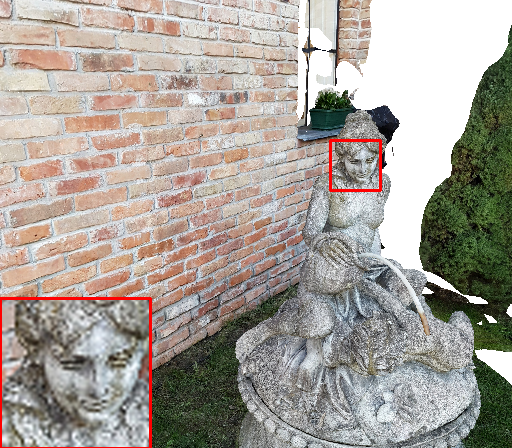}
      Mesh Ground Truth
  \end{subfigure}
  \begin{subfigure}{0.3\linewidth}
    \centering
        \includegraphics[width=0.9\linewidth]{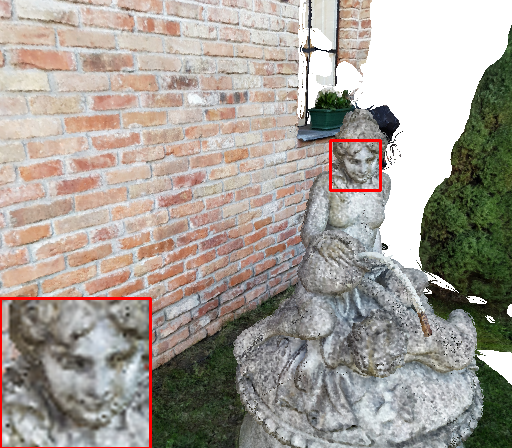}
        Pointersect~\cite{chang2023pointersect}
    \end{subfigure}
    \vspace{2mm}

    \begin{subfigure}{0.3\linewidth}
    \centering
        \includegraphics[width=0.9\linewidth]{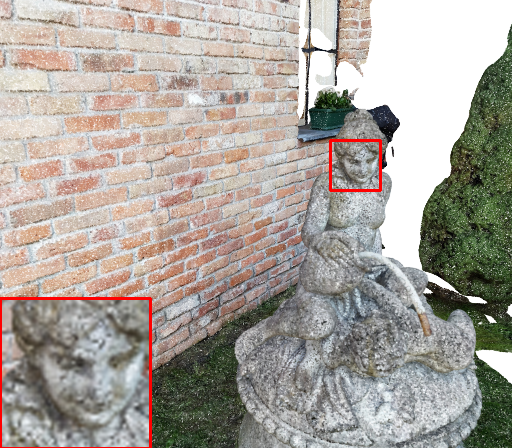}
        Global Parameter
    \end{subfigure}
    \begin{subfigure}{0.3\linewidth}
    \centering
        \includegraphics[width=0.9\linewidth]{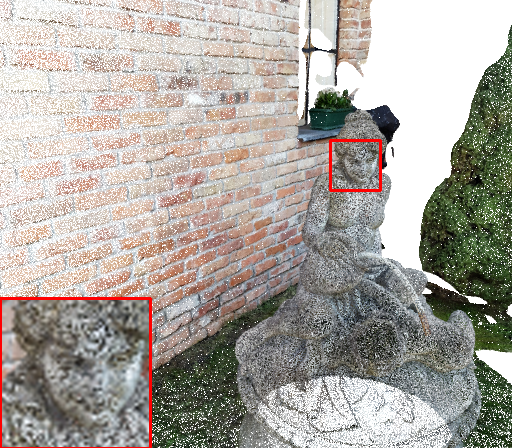}
        OpenGL
    \end{subfigure}
    \begin{subfigure}{0.3\linewidth}
    \centering
        \includegraphics[width=0.9\linewidth]{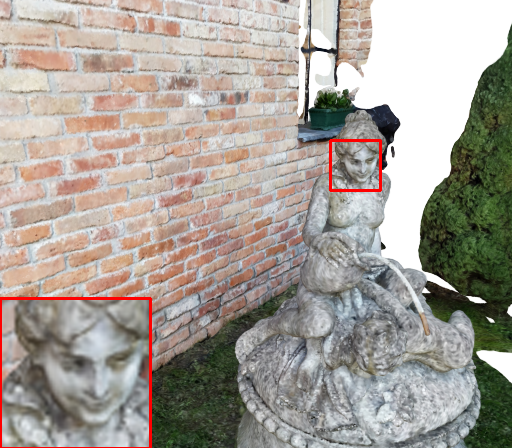}
        Ours
    \end{subfigure}
    \vspace{2mm}

    \begin{subfigure}{0.3\linewidth}
      \centering
        \includegraphics[width=0.9\linewidth]{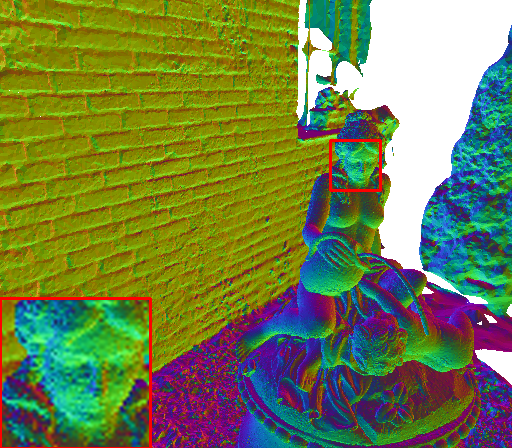}
        Mesh Ground Truth
    \end{subfigure}
    \begin{subfigure}{0.3\linewidth}
      \centering
          \includegraphics[width=0.9\linewidth]{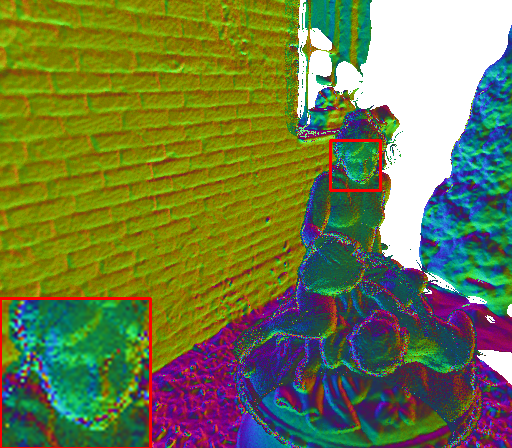}
          Pointersect~\cite{chang2023pointersect}
      \end{subfigure}
    \vspace{2mm}
    
      \begin{subfigure}{0.3\linewidth}
      \centering
          \includegraphics[width=0.9\linewidth]{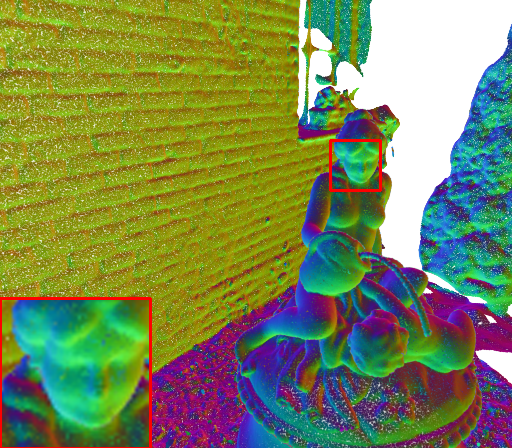}
          Global Parameter
      \end{subfigure}
      \begin{subfigure}{0.3\linewidth}
      \centering
          \includegraphics[width=0.9\linewidth]{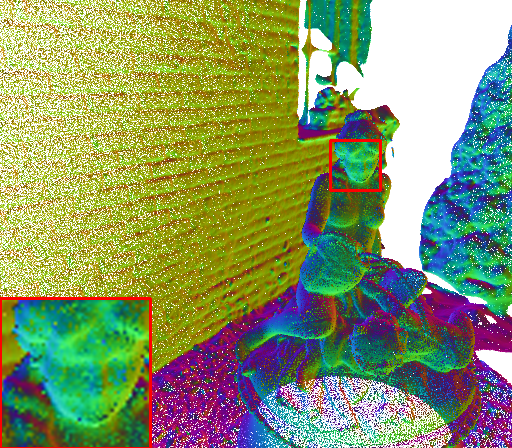}
          OpenGL
      \end{subfigure}
      \begin{subfigure}{0.3\linewidth}
      \centering
          \includegraphics[width=0.9\linewidth]{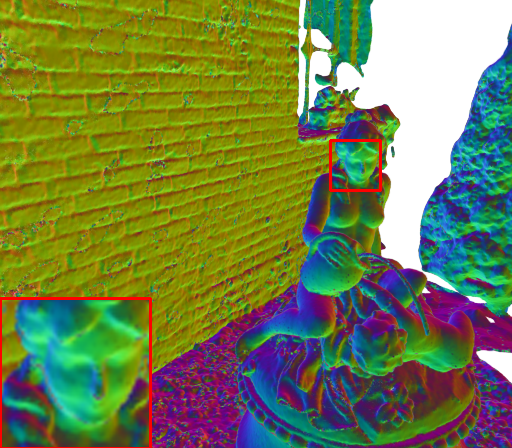}
          Ours
      \end{subfigure}
  
  \caption{Rendering results of high quality point clouds in BlendedMVS dataset. The insets visualize local details with $3\times$ zooming.}
  \label{fig:1032}
\end{figure*}

\begin{figure*}[t]
  \centering

  \begin{subfigure}{0.3\linewidth}
  \centering
      \includegraphics[width=\linewidth]{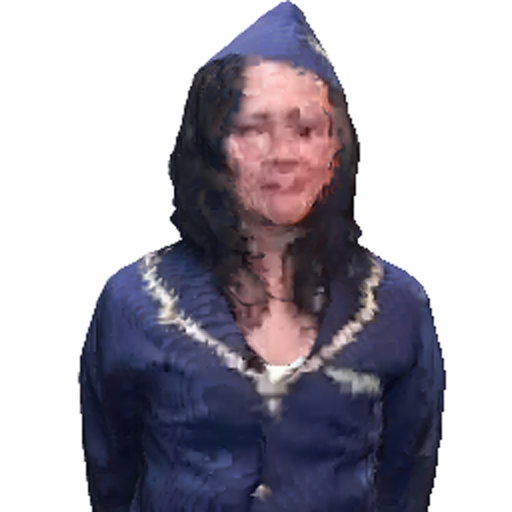}
      Poisson
  \end{subfigure}
  \begin{subfigure}{0.3\linewidth}
    \centering
        \includegraphics[width=\linewidth]{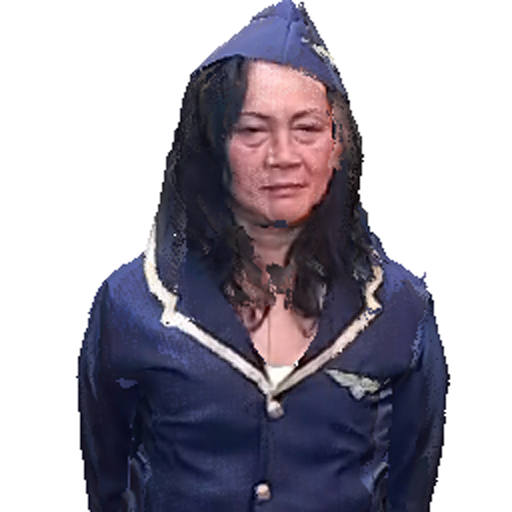}
        Pointersect~\cite{chang2023pointersect}
    \end{subfigure}

  \begin{subfigure}{0.3\linewidth}
    \centering
        \includegraphics[width=\linewidth]{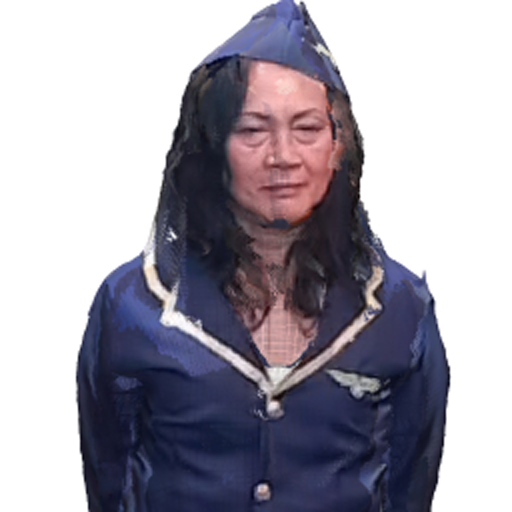}
        Global Parameter
    \end{subfigure}
  \begin{subfigure}{0.3\linewidth}
  \centering
      \includegraphics[width=\linewidth]{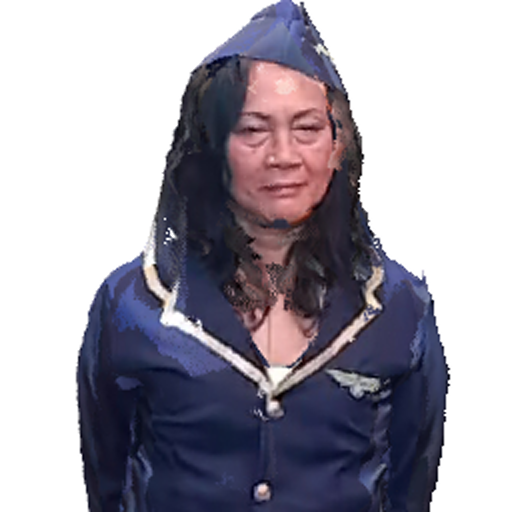}
      OpenGL
  \end{subfigure}
  \begin{subfigure}{0.3\linewidth}
  \centering
      \includegraphics[width=\linewidth]{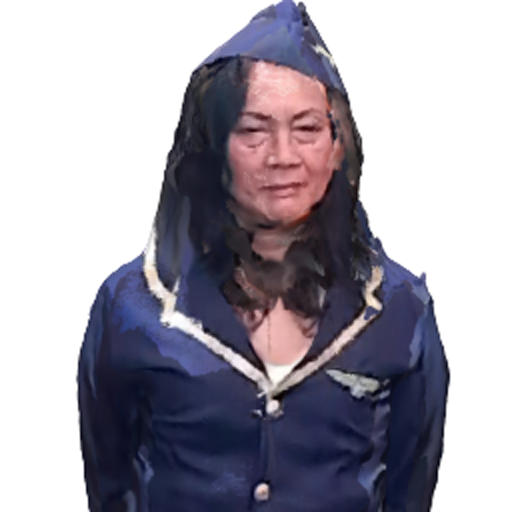}
      Ours
  \end{subfigure}
  
  \caption{Rendering results from noisy raw point clouds in the CWIPC dataset.}
  \label{fig:cwis3}
\end{figure*}

We visualize more rendering results with high-quality (800K points) point clouds from the THuman 2.0 dataset in \Cref{fig:0513_800k} and \ref{fig:0519_800k}, outdoor scenes (1.5M points) from the BlendedMVS dataset in \Cref{fig:1032}, and noisy raw captures (1M points) from the CWIPC dataset in  
\Cref{fig:cwi} and \Cref{fig:cwis3}.

We also evaluate our method in dynamic quality by rendering the point cloud video sequences from the 8iVFB database~\cite{dataset8i}. We use a video framebuffer at the resolution of $1024 \times 1024$ and the frame rate at 30 fps, in accordance to the original point cloud framerate. Please refer to the supplementary video for the results. As shown, our method can render high-quality videos without visible holes which are observed with the standard OpenGL renderer. Therefore, our method is more suitable for rendering high-quality point cloud videos in real-time applications.

\begin{figure*}
    \centering
    
    \begin{subfigure}{0.48\linewidth}
        \centering
        \includegraphics[width=0.9\linewidth]{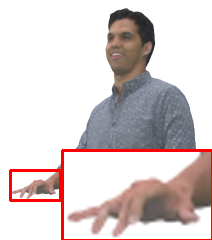}
        Ours
    \end{subfigure}
    \begin{subfigure}{0.48\linewidth}
        \centering
        \includegraphics[width=0.9\linewidth]{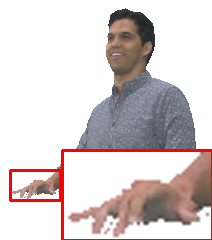}
        Pointersect~\cite{chang2023pointersect}
    \end{subfigure}
    \begin{subfigure}{0.48\linewidth}
        \centering
        \includegraphics[width=0.9\linewidth]{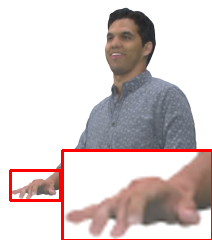}
        Global Parameter
    \end{subfigure}
    \begin{subfigure}{0.48\linewidth}
        \centering
        \includegraphics[width=0.9\linewidth]{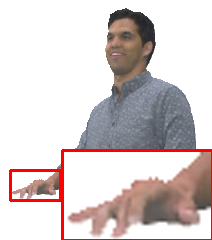}
        OpenGL
    \end{subfigure}

    \caption{Rendering results of different methods on the sequence \textit{Loot} (610 K points, 23 Mbps) in the 8iVFB database. The insets visualize local details with $3\times$ zooming.}
    \label{fig:gpcc_loot}
    \vspace{-2mm}
\end{figure*}

\begin{figure*}[t]
    \centering


    \begin{subfigure}{0.3\linewidth}
    \centering
        \includegraphics[width=\linewidth]{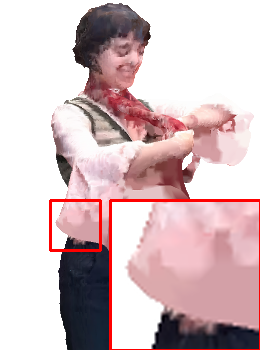}
        Poisson
    \end{subfigure}
    \begin{subfigure}{0.3\linewidth}
    \centering
        \includegraphics[width=\linewidth]{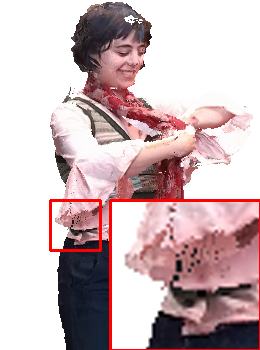}
        Pointersect~\cite{chang2023pointersect}
        \end{subfigure}

    \begin{subfigure}{0.3\linewidth}
    \centering
        \includegraphics[width=\linewidth]{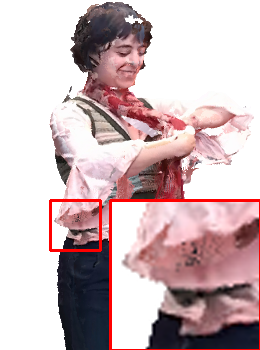}
        OpenGL
    \end{subfigure}
    \begin{subfigure}{0.3\linewidth}
    \centering
        \includegraphics[width=\linewidth]{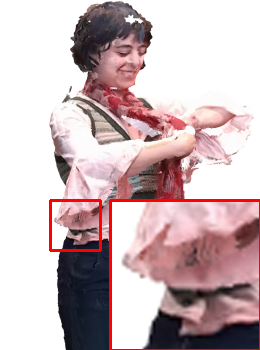}
        Global Parameters
    \end{subfigure}
    \begin{subfigure}{0.3\linewidth}
    \centering
        \includegraphics[width=\linewidth]{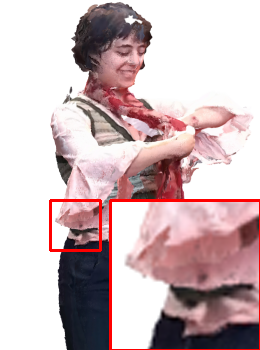}
        Ours
    \end{subfigure}
    
    \caption{Rendering results from noisy raw point clouds in the CWIPC dataset. The insets visualize local details with $2\times$ and $3\times$ zooming, respectively.}
    \label{fig:cwi}
\end{figure*}

\end{document}